\pdfoutput=1

\documentclass[11pt]{article}

\usepackage[preprint]{acl}

\usepackage{times}
\usepackage{latexsym}

\usepackage[T1]{fontenc}

\usepackage[utf8]{inputenc}

\usepackage{microtype}

\usepackage{inconsolata}

\newcommand\blfootnote[1]{%
  \begingroup
  \renewcommand\thefootnote{}\footnote{#1}%
  \addtocounter{footnote}{-1}%
  \endgroup
}

\usepackage{hyperref}
\usepackage{url}
\usepackage{mkolar_definitions}
\usepackage{booktabs}
\usepackage{makecell}
\usepackage{multirow}
\usepackage{graphicx}
\usepackage{xspace}
\usepackage{enumitem}
\usepackage[normalem]{ulem}

\newcommand{\rotatedMultiRow}[2]{\parbox[t]{2mm}{\multirow{#2}{*}{\rotatebox[origin=c]{90}{#1}}}}

\definecolor{darkgreen}{rgb}{0,0.5,0}
\definecolor{lightgreen}{rgb}{0.63,0.89,0.72}
\definecolor{darkred}{rgb}{0.7,0,0}
\definecolor{tealtwo}{rgb}{0.1,0.6,0.7}
\definecolor{blue}{rgb}{0.0,0.1,0.9}
\definecolor{yellow}{rgb}{0.98,0.95,0.1}
\definecolor{orange}{rgb}{1.,0.7,0.0}
\definecolor{lightblue}{rgb}{0.70, 0.80, 0.89}
\definecolor{lightred}{RGB}{214,39,40}
\definecolor{darkblue2}{RGB}{31,119,180}
\definecolor{darkyellow}{RGB}{160,161,36}
\definecolor{mypink2}{RGB}{219, 48, 122}

\title{Analysis of Plan-based Retrieval for Grounded Text Generation}

\author{Ameya Godbole$^{\diamondsuit \dagger *}$, Nicholas Monath$^{\spadesuit *}$, Seungyeon Kim$^{\clubsuit}$, Ankit Singh Rawat$^{\clubsuit}$ \\
 {\bf Andrew McCallum$^{\spadesuit}$, Manzil Zaheer$^{\spadesuit}$} \\
  $^{\diamondsuit}$University of Southern California \qquad $^{\spadesuit}$Google DeepMind \qquad $^{\clubsuit}$Google Research \\
  \texttt{ameyagod@usc.edu} \hspace{5pt} \texttt{nmonath@google.com} \\
  }

\begin{document}

\maketitle

\begin{abstract}
In text generation, hallucinations refer to the generation of seemingly coherent text that contradicts established knowledge. One compelling hypothesis is that hallucinations occur when a language model is given a generation task outside its parametric knowledge (due to rarity, recency, domain, etc.). A common strategy to address this limitation is to infuse the language models with retrieval mechanisms, providing the model with relevant knowledge for the task. In this paper, we leverage the planning capabilities of instruction-tuned LLMs and analyze how planning can be used to guide retrieval to further reduce the frequency of hallucinations. We empirically evaluate several variations of our proposed approach on long-form text generation tasks. By improving the coverage of relevant facts, plan-guided retrieval and generation can produce more informative responses while providing a higher rate of attribution to source documents.
\end{abstract}

\section{Introduction}

\blfootnote{$^\dagger$Work done as a Student Researcher at Google.}
\blfootnote{$^*$Equal contribution.}

Large, parametric language models (LLMs) provide 
highly fluent text for many applications such as 
summarization, dialogue, and translation \citep[inter alia]{devlin2018bert,brown2020language,thoppilan2022lamda,chowdhery2022palm,palm2}. Among the errors made by these models, producing generations with factual and/or grounding errors, often referred to as \emph{hallucinations}, limit the broader applicability and capability of language models \citep{gao2022attributed,manakul2023selfcheckgpt,min2023factscore,ji2023survey,peng2023check}. Hallucinations differ from other kinds of errors in that the generated text is syntactically correct and semantically plausible. These hallucinations are generations that, 
were they factually accurate, would be satisfactory model output.

As an example of such hallucinations, we consider the following generation from the Falcon 180B model demo\footnote{https://huggingface.co/spaces/tiiuae/falcon-180b-demo} for the query \emph{write a bio about the author lorrie moore}. 

\begin{quote}
{{\footnotesize
    lorrie moore is an american writer {\bf \textcolor{darkred}{born in glasgow, kentucky}} in 1957. she graduated from saint lawrence university in canton, new york in 1978, and earned her mfa from the \textcolor{darkred}{\bf university of wisconsin-madison} in 1982. moore has written numerous works of fiction, including short story collections such as self help, like life, and birds of america, as well as novels like anagrams, who will run the frog hospital?, and a gate at the stairs. moore has won many awards, including \textcolor{darkred}{\bf the national book critics circle award for fiction in 1990}, the pen/malamud award for excellence in short fiction in \textcolor{darkred}{\bf 1994}, and the rea award for the short story in \textcolor{darkred}{\bf 2018}. her writing style is often described as sharp, funny, and introspective, with themes centered around family relationships, love, loss, and the complexities of modern life.}}
\end{quote}
While the model accurately generates the book titles, it hallucinates many facts \textbf{\textcolor{darkred}{(red, bold-faced)}} about Moore such as the place of her birth (she was born in Glens Falls, New York), the university of her MFA (she received an MFA from Cornell University). The model also confuses all of the awards given to the author (incorrect years, award names, missing awards). 

Recent work on detecting and mitigating such hallucinations has included modeling uncertainty \citep{lee2022factuality}, constrained decoding \citep{lu2021neurologic}, specially trained or auxiliary model-based decoding \citep{zhou2023controlled,roit2023factually}, question-answering planning \citep{narayan2022conditional,huot2023text}, and retrieval-based verification of claims \citep{gao2023rarr}. 
These methods demonstrate success by leveraging the model's ability to accurately produce factual information via its parameters or through text provided in-context. However, understanding and analyzing {where} and {why} such hallucinations appear and {why} providing evidence in-context can successfully mitigate hallucinations has only begun to be studied recently \cite{li2023halueval,das2023diving,sadat2023delucionqa}.

In this work, we study how planning can be used  to guide retrieval to improve factual text generation.
We hypothesize that language models hallucinate when they are required to generate certain facts based on the prompt but do not have the information either memorized or in their context.
Thus, we hypothesize that with specific, comprehensive 
facts in-context can mitigate hallucinations.
Our investigation considers the following questions:
\vspace{1mm}
    
\noindent\textbf{Q1.} How to effectively discover the comprehensive collection of facts needed to generate text about a particular subject? (\S\ref{sec:retrieval},\S\ref{sec:planning}, Table~\ref{tab:main_table_bison}, \ref{tab:main_table_unicorn}, \ref{tab:mistral_7b})

\noindent\textbf{Finding:} We observe that generating search queries based on the LLM plans lets the retrieval system gather fine-grained facts that the LLM needs to write its final response.

\noindent\textbf{Q2.} How can we effectively retrieve and represent these needed facts? (\S\ref{sec:planning}, \S\ref{sec:method_variants}, Table~\ref{tab:second_search_ablation}, \ref{tab:unanswerable} \ref{tab:outlingImportance})

\noindent\textbf{Finding:} We observe that the groundedness of the final model response is heavily influenced by the retrieved information and not as much by the wording of the prompts. We also show that including unanswerable search queries (and explicitly marking them as unanswerable) is influential in reducing the generation of ungrounded outputs.

\vspace{1mm}
    
We demonstrate the generality of the approach in writing grounded text by evaluating model responses in two domains: writing biographies and writing event descriptions. In addition to entities and events that may have a significant internet presence, we specifically consider 2 settings that would challenge LLMs: (1) writing biographies for people (researchers) who do not have Wikipedia profiles (long-tail, low-frequency entities) and (2) writing about current news events (outside of the parametric knowledge of the LLMs).

\begin{figure*}[t]
    \vspace{-5mm}
    \centering
    \includegraphics[width=\textwidth]{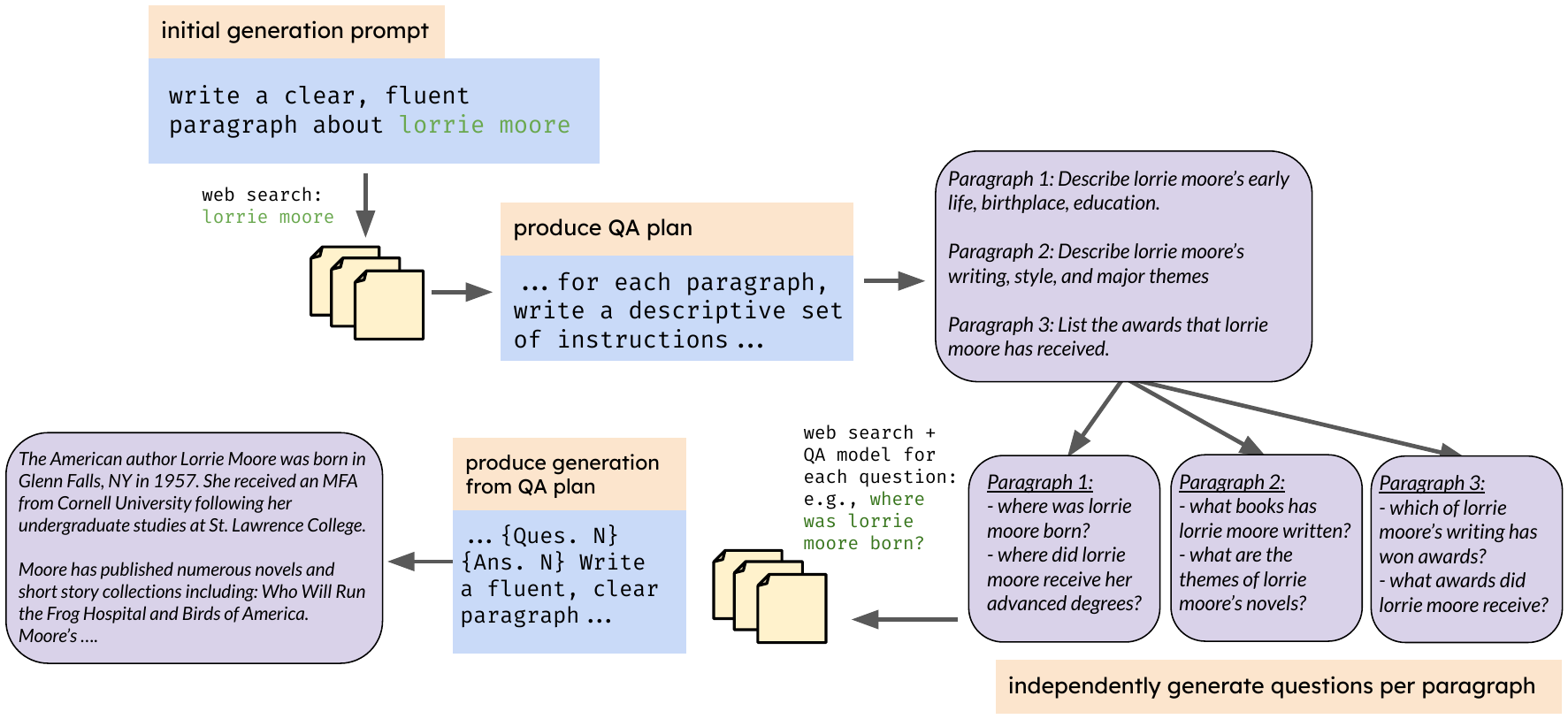}
    \caption{\textbf{Summary of Planning and Retrieval used to generate text.} Given an initial prompt, a plan is first generated that outlines the segments to be written. Next, search queries are generated for each segment which are then used for fine-grained retrieval retrieval of source documents. The final response is generated conditioned on the plan, the queries and the retrieved documents.
    }
    \label{fig:planning}
        \vspace{-5mm}
\end{figure*}

\section{Related Work}
\label{sec:related_work}

The focus of our paper is to provide analysis as to how retrieval augmented language models can reduce hallucinations by conditioning on relevant facts in-context. 
This work abuts many related areas in text generation (factuality-focused and otherwise), in-context learning, retrieval, and more. Refer to other related work in App~\ref{app:ext_related_work}.

\noindent\textbf{Verifying Attribution.}  \citet{rashkin2023measuring} introduce formal definitions of what it means for a piece of text to be attributable to a given source, referred to as \emph{AIS} (Attributable to Identified Sources). This definition has been used to design datasets and models to automatically predict whether a particular source document supports a claim. Verification models have considered a variety of formulations such as question-answering based \citep{honovich2021q}, lexical alignment based \citep{goyal-durrett-2020-evaluating} and NLI-based \citep{honovich2022true,gekhman2023trueteacher}.
Among these, NLI-based models are a commonly used automated AIS metrics to measure whether a particular output is grounded in provided sources
\citep[inter alia]{bohnet2022attributed,gao2023rarr}.

By incorporating web or corpus search, the AIS models can be used to check whether a particular claim is true with respect to general world knowledge \cite{min2023factscore,peng2023check}.
FactScore \citep{min2023factscore} prompts an LLM to break down claims into atomic facts to be verified.
We use a simplified version of the FactScore metric by compute sentence-level entailment scores. These search-enabled verification models have enabled the study of how language models can generate attributed text along with citations and study accuracy of such deployed approaches \citep{liu2023evaluating,gao2023enabling}.

\noindent\textbf{Retrieval for Generating Attributable Text.} It is well known that language models store a large amount of knowledge in their parameters \citep{petroni2019language,petroni2020context}. Retrieving evidence documents allows LLMs to synthesize text about facts outside their parametric knowledge~\citep{mallen2022not}, including using up-to-date~\citep{zhu2020modifying,vu2023freshllms} and proprietary~\citep{min2023silo} information. Past work has incorporated retrieval for post-hoc verification and editing~\citep{gao2023rarr,chen2023purr}. Other approaches incorporate search results throughout the generation process either based on heuristics~\citep[inter alia]{trivedi-etal-2023-interleaving,amplayo-etal-2023-query,jiang-etal-2023-active,press-etal-2023-measuring} or based on trigger tokens generated by the model~\cite{asai2023selfrag}.

\noindent\textbf{Planning of Long-form Generation.} Past work has demonstrated that planning (outlining the output) before generation can help to improve the factuality and quality of the model output. QA blueprint-based methods \citep{narayan2022conditional,huot2023text} use question-answer pairs to change the LLM distribution over output text to make summarization models more grounded. \citet{huot2024muplan} extend this approach to cross-lingual summarization. \citet{akash2023longform} use a trained model to generate a keyword plan before generating a each new sentence and use the plan to inform search. \textbf{Our contribution:} We study whether zero-shot LLM generated plans can be used to retrieve relevant information beyond simple search and how the plans along with the new, diverse search results can guide the final generation to be more grounded. Concurrently with this work, \citet{shao2024assisting-storm} also demonstrate the importance of pre-writing (outlining, iterative search and outline refinement) for long-form expository writing, further supporting the findings of our work.

\section{Empirical Study}
\label{003_empirical_methods}

Our goal is to understand the effect of providing 
more complete contextual information to 
language models on hallucination in text generation.
We compare different strategies for performing retrieval and placing facts into context.

We consider the task of generating descriptive text about an {\tt entity}. This task effectively constitutes writing a bio about an individual (such as {\tt author Lorrie Moore} in the previous example) or a summary about an event. This text will then be verified against web sources for attribution using automated methods~\citep{honovich2022true}.

\subsection{Direct Generation without Retrieval}
\label{sec:pureparametric}

The simplest approach is to directly prompt the LLM
to generate text about the given \texttt{entity}. This approach
relies on information known by the parametric language model without any additional retrieval. We provide examples of the prompts used in \S\ref{app:prompts}.1, which use entity name and possibly a disambiguating attribute (occupation, location, etc.).

\subsection{Incorporating Retrieval}
\label{sec:retrieval}

Towards providing the relevant facts needed for the LLM to write about a particular \texttt{entity}, we simply perform web search with the \texttt{entity} name as the query and provide the search results (titles and text) in the model context to allow retrieval augmented generation \citep{gao2023enabling} (see \S~\ref{app:prompts}.2).

\subsection{Planning and Blueprints}
\label{sec:planning}

Past work has demonstrated that LLMs can generate reasoning chains to improve their task performance \citep[interalia]{wei2022chain}. Additionally,  \citet{narayan2022conditional} have shown that question-answer based plans allow models to generate more faithful text. We combine these abilities of LLMs to guide retrieval and provide the information that the model may need to generate a response without (or with fewer) hallucinations. 

We first prompt the model to write a list describing what paragraphs should be produced (see \S\ref{app:prompts}.3). Each item in this outline provides short sentence or two about the content for each paragraph. The outline is conditioned on the initial search results (using a query of the \texttt{entity} name) to guide the LLM outline to generate targeted, entity-specific outlines rather than generic plans. 

Next, we generate search queries from the paragraph descriptions. 
We take the content description for each paragraph and the initial search results as context and generate web search queries (see \S\ref{app:prompts}.4). These queries allow the search engine to gather information about specific aspects of the \texttt{entity}. The result of this prompt is a list of questions such as those shown in Figure~\ref{fig:planning} for each paragraph. 
These questions will be used to gather additional
web search results.

We consider two variants for incorporating these 
newly gathered web-search results. The first, 
simply places the new results in context. The second 
uses a separate off-the-shelf span-selection QA model to provide answers for each question. We use a confidence threshold to decide how many answers to include in the prompt for the final generation. If there are no answers that the QA model is confident about, we keep the question but mark the answer as `\texttt{unanswerable}'. We combine the questions for all paragraphs together to form the prompt we use to generate the final output (see \S\ref{app:prompts}.5).

\newcommand{\textbison}{text-bison-001\xspace}
\newcommand{\textunicorn}{text-unicorn-001\xspace}

\newcommand{\paramOnly}{No Retrieval\xspace}
\newcommand{\searchOnly}{One-Retrieval\xspace}
\newcommand{\searchOnlyLong}{One-Retrieval (2x snippets)\xspace}
\newcommand{\planAndRetrieve}{Plan-based Retrieval (Var.B)\xspace}
\newcommand{\planAndRetrieveSprompt}{w/ Alt. Prompt}
\newcommand{\planTheSearch}{Plan-based Retrieval (Var.A)\xspace}

\begin{table*}
    \begin{center}\footnotesize
    \begin{tabular}{l | l | rrr | rrr | r}
    \toprule
    & \bf Approach & \multicolumn{3}{c|}{\bf AIS} & \multicolumn{3}{c|}{\bf Rouge Prec.} & \bf Length \\
     &  & Strict & Macro & Micro & R1 & R2 & RL & \# Tokens \\ 
     \midrule
     \midrule
     \rotatedMultiRow{Wiki-Event}{5} & \paramOnly & 0.00 & 15.30 & 15.74 & 69.45 & 30.16 & 66.63 & 88.63  \\
     & \searchOnly & 74.54 & 90.46 & 90.35 & 96.23 & 84.94 & 95.09 & 79.26\\
     & \searchOnlyLong & 75.39 & 90.99 & 91.10 & 97.65 & 87.01 & 97.04 & 78.63 \\
     & \planTheSearch &   \bf89.99 & \bf 96.18 & \bf 95.14 & \bf 99.59 & \bf 94.52 & \bf 99.44 & 82.76 \\
     & \planAndRetrieve & 72.96 & 92.02 & 91.85 & 99.43 & 90.30 & 99.19 & 113.92 \\
     \midrule
     \rotatedMultiRow{Wiki-Ent}{5} & \paramOnly  & 0.00 & 13.80 & 16.27 & 55.67 & 16.95 & 53.14 & 97.29 \\
     & \searchOnly  & 60.91 & 83.09 & 83.11 & 91.52 & 74.17 & 90.12 & 87.62 \\
     & \searchOnlyLong & 67.03 & 87.44 & 86.67 & 95.79 & 80.45 & 94.92 & 86.86 \\
      & \planTheSearch  & \bf 84.77 &\bf  95.16 &\bf  94.23 &\bf  99.59 & \bf 92.99 &\bf  99.44 & 93.03 \\
      & \planAndRetrieve  & 63.37 & 90.40 & 90.70 & 99.28 & 89.45 & 99.04 & 134.47 \\
      \midrule
     \rotatedMultiRow{Researcher}{5} & \paramOnly & 0.00 & 6.74 & 6.73 & 57.10 & 17.85 & 54.84 & 80.58  \\
     & \searchOnly  & 62.89 & 83.68 & 84.70 & 91.55 & 73.90 & 90.53 & 74.24 \\
     & \searchOnlyLong & 69.81 & 86.34 & 87.57 & 95.90 & 80.03 & 95.27 & 75.01 \\
      & \planTheSearch &\bf  79.56 &\bf  91.63 &\bf  92.16 & \bf 97.87 & \bf 83.99 & \bf 97.48 & 77.71 \\
      & \planAndRetrieve  &  64.78 & 88.31 & 88.93 & 97.72 & 83.40 & 97.23 & 98.00\\
     \midrule
      \rotatedMultiRow{News Events}{5} & \paramOnly & 0.00 & 5.46 & 5.14 & 68.81 & 24.41 & 66.00 & 89.44  \\
         & \searchOnly & 64.10 & 84.76 & 87.77 & 95.95 & 80.93 & 94.61 & 83.51 \\
         & \searchOnlyLong & 61.54 & 86.72 & 87.62 & 97.63 & 80.70 & 96.50 & 79.83 \\
        & \planTheSearch & \bf 80.77 & \bf 94.47 &\bf  93.72 & \bf 99.76 & \bf 93.47 & \bf 99.60 & 92.15 \\
        & \planAndRetrieve & 66.67 & 90.88 & 91.49 & 99.61 & 92.53 & 99.46 & 124.67 \\
         \bottomrule
    \end{tabular}
    \end{center}
       \caption{\textbf{Comparison of Generation Approaches using \textbison model.} The plan-based retrieval models that are at the center of our analysis yield more attributable text than One- and No- Retrieval methods. Of the two variants, Var.A produces more attributable sentences and Var.B produces longer texts.}
    \label{tab:main_table_bison}
    \vspace{-5mm}
\end{table*}
\vspace{-1mm}
\section{Empirical Analysis}
\vspace{-1mm}

We will measure quantitatively the ability of models 
to provide attributed generations. 
We will analyze how producing attributed text 
is affected by the way in which evidence is retrieved and used. 
In particular, we aim to investigate the following: 
\vspace{1mm}

    
    
    
    
    \noindent \textbf{Q1}: Does providing more comprehensive collections of facts yield more attributable generations?
    
    \textbf{\S\ref{exp:results} Observations:} 1-3
    
    \noindent \textbf{Q2}: Can planning and additional web-searches improve the collection of facts used for generations?
    
    \textbf{\S\ref{exp:results} Observations:} 4, 7, 8
    
    \noindent \textbf{Q3}: Which representations of facts in-context are most effective for producing grounded texts?
    
    \textbf{\S\ref{exp:results} Observations:} 5, 6

\subsection{Datasets}
\label{sec:dataset}
As described in \S\ref{sec:pureparametric}, the task we consider is writing a biography of an entity or summary of an event. We select this task because it likely necessitates gathering of knowledge more so than tasks such as summarization in which most (if not all) of the required information is provided. These datasets used for the task are meant to cover both entities and events from the head and the tail of the parametric knowledge of LLMs. 

The setting we consider is one where the system is presented only with an entity or event name and (possibly) an adjective for entity disambiguation (e.g., {\tt Gerhard Fischer (inventor)}). Since our evaluation focus is about hallucination,  we do not consider defining a ``ground-truth'' biography or event summary, but rather determine if the generated text is grounded. We collect entity and event names using the following:

    \noindent \textbf{Researcher}. We consider writing bios for the organizing boards and committee for: NeurIPS 2023, 2023 organizing committee for ASTMH (American Society for Tropical Medicine and Hygiene), the 27th Nordic Particle Physics Meeting (2023), and the American Comparitive Literature Association (ACLA). This results in a list of 106 entity names. We observe that there is sufficient information on the internet to write a biography for these entities, yet most do not have a Wikipedia page. We hypothesize that the entities in this dataset are likely in the `tail' of the distribution of entity mentions. 

    \noindent \textbf{Wiki-Ent}. To select a collection of entities closer to the `head' of the distribution of entity mentions, we select a subset of 81 entities from the list of entities used by \citep{min2023factscore}. Each entity has a Wikipedia page (though we do not treat the Wikipedia page in any special way, it may or may not be retrieved by search-based methods). 
    
    \noindent \textbf{Wiki-Event}. We collect a list of 233 events from the validation set of the WEC corpus \citep{eirew-etal-2021-wec}. These events that are the target for coreference resolution in the original dataset. Similar to the Wiki-Ent collection, these event names have corresponding Wikipedia pages (we neither forcibly include nor exclude that page from search).
    
    \noindent \textbf{News Events}. We select a list of 52 News Events that began in August 2023\footnote{\url{https://en.wikipedia.org/wiki/Portal:Current_events/August_2023}}. These events are chosen for their recency, i.e. they occurred after the knowledge cutoff of the LLM.

\subsection{Methods Compared}
\label{sec:method_variants}

We evaluate two models of different sizes from
Google Cloud VertexAI\footnote{\url{https://cloud.google.com/vertex-ai/docs/generative-ai/learn/model-versioning}},
namely
(the smaller) \textbison and (the larger) \textunicorn \citep{palm2}
with zero-shot instructions following the templates as described in \S\ref{003_empirical_methods}. Additionally, we report the effectiveness of grounded generation with Plan-based Retrieval on the open-weight Mistral-7B-Instruct model in \S~\ref{sec:mistral_7b}.

We compare the following retrieval methods: 

    \noindent \textbf{\paramOnly}: Baseline approach described in \S~\ref{sec:pureparametric} where the language model prompted to use it's parametric knowledge to directly generate the output without any external context.
    
    \noindent \textbf{\searchOnly}: This is the approach described in \S~\ref{sec:retrieval} which uses one round of web search and concatenates the results in-context to the model. This formulation provides the language model external knowledge for conditional text generation.  
    
    \noindent \textbf{\searchOnlyLong}: This approach uses a single round of retrieval, but uses double the number of search results compared to \searchOnly. Note this is also double the number of search results used by the plan-based methods in their first round of search as well.
    
    \noindent \textbf{Plan-based Retrieval}: The proposed approach that uses planning (\S~\ref{sec:planning}) to retrieve additional web search results. There are two variants, which differ in how the additional plan-based retrieved information is used to write the final response: i) Variant A simply puts the web search results in-context, and ii) Variant B performs the aforementioned QA-based outline. We further describe and compare several ablations of Plan-based Retrieval in \S\ref{exp:results}.

\begin{table*}
    \begin{center}\footnotesize
    \begin{tabular}{l | l | rrr | rrr | r}
    \toprule
     & \bf Approach & \multicolumn{3}{c|}{\bf AIS} & \multicolumn{3}{c|}{\bf Rouge Prec.} & \bf Length \\
     &  & Strict & Macro & Micro & R1 & R2 & RL & \# Tokens \\ 
    \midrule
    \midrule
        \rotatedMultiRow{Wiki-Event}{5}
         & \paramOnly &  1.00 & 25.80 & 23.70 & 70.39 & 33.34 & 67.82 & 135.91 \\
         & \searchOnly & 68.67 & 89.39 & 88.47 & 95.58 & 85.77 & 94.71 & 98.76 \\
         & \searchOnlyLong & 67.53 & 88.32 & 85.96 & 96.74 & 86.65 & 95.99 & 100.90 \\
          &\planTheSearch & \bf  88.27 & \bf 96.33 & \bf 96.25 & \bf 99.66 & \bf 94.45 & \bf 99.52 & 98.61\\
         & \planAndRetrieve & 82.40 & 94.50 & 94.13 & 99.36 & 92.82 & 99.19 & 115.45 \\
         \midrule
          \rotatedMultiRow{Wiki-Ent}{5} 
         & \paramOnly &  0.00 & 17.79 & 20.31 & 56.11 & 19.00 & 53.53 & 115.45  \\
         & \searchOnly & 63.79 & 86.70 & 87.37 & 92.28 & 74.72 & 90.87 & 87.30 \\
         & \searchOnlyLong & 58.65 & 85.67 & 83.96 & 94.46 & 79.43 & 93.42 & 110.23 \\
          &\planTheSearch & \bf 76.54 & \bf 94.31 & \bf 94.25 & \bf 99.41 & \bf 91.81 & \bf 99.26 & 127.22\\
          & \planAndRetrieve & 69.96 & 93.58 & 93.02 & 99.28 & 90.96 & 99.06 & 164.23 \\
         \midrule
          \rotatedMultiRow{Researcher}{5}
         & \paramOnly &  0.00 & 7.90 & 7.82 & 55.38 & 15.96 & 52.73 & 99.20  \\
         & \searchOnly & 62.26 & 86.38 & 87.36 & 91.95 & 75.05 & 90.99 & 78.82 \\
         & \searchOnlyLong & 63.21 & 85.09 & 86.04 & 95.91 & 82.08 & 95.26 & 93.65 \\
         & \planTheSearch & \bf 85.22 & \bf 94.31 & \bf 94.68 & \bf 98.33 & \bf 87.74 & \bf 98.07 & 87.73\\
          & \planAndRetrieve & 81.76 & 93.29 & 93.81 & 98.12 & 86.86 & 97.84 & 104.62 \\
         \midrule
          \rotatedMultiRow{News Events}{5}
         & \paramOnly & 1.92 & 12.69 & 9.89 & 68.91 & 25.33 & 66.36 & 120.89  \\
         & \searchOnly & 67.31 & 90.27 & 88.21 & 95.83 & 83.48 & 94.52 & 96.42   \\
         & \searchOnlyLong & 63.46 & 88.42 & 85.79 & 96.37 & 82.81 & 95.46 & 104.11 \\
         & \planTheSearch & \bf 87.18 & \bf 96.86 & \bf 96.68 & \bf 99.73 & \bf 94.05 & \bf 99.62 & 100.91 \\
         & \planAndRetrieve & 82.69 & 95.85 & 96.01 & 99.72 & 93.62 & 99.55 & 135.72 \\
         \bottomrule
    \end{tabular}
    \end{center}
        \caption{\textbf{Comparison of Generation Approaches using \textunicorn model.} We observe that Plan-based Retrieval improves upon One-Retrieval and No Retrieval methods, even when One-Retrieval retrieves more results. Plan-based Retrieval Var.B produces much longer texts, which are more attributable compared to One-Retrieval. Var.A produces slightly more attributed texts and at shorter length.}
    \label{tab:main_table_unicorn}
\end{table*}

\begin{table}[]
    \begin{center}
    \footnotesize
    \begin{tabular}{@{}l@{}r@{}r@{}r@{}r@{}r@{}r@{}}
    \toprule
    \bf  & \multicolumn{3}{c}{\bf AIS} & \multicolumn{2}{c}{\bf Rouge P.} & \bf {Len.} \\
     &  \ Strict & Macro& \ Micro  & \ R2 &\  RL & \# Tok \\ \midrule\midrule
    One-Retr. &  $\ $ 60.91  &  $\ $ 83.09 & $\ $  83.11 & $\ $  74.17 & $\ $  90.12 & $\ $  87.62 \\
     \midrule
      Plan-based (Var.B) \ &  $\ $ 63.37  &  $\ $  90.40 &  $\ $  90.70   & $\ $  89.45 & $\ $  99.04 & $\ $  134.47 \\
        {\scriptsize w/o 2nd search \ }  &  $\ $ 61.32  & $\ $ 85.46    &    $\ $85.13  &   $\ $ 80.05 &  $\ $  94.11 &   $\ $ 98.72 \\
        \bottomrule
    \end{tabular}
    \end{center}
        \caption{\textbf{Importance of Gathering Additional Information from Second Search.} We compare \planAndRetrieve{} with \textbison in two settings, where we use the typical (secondary retrieval step) and using only the original search results as the source for the documents. We measure performance on the Wiki-Ent dataset. We find that attribution is indeed improved by gathering more facts and information.}
    \label{tab:second_search_ablation}
\end{table}

\begin{table*}[]
    \begin{center}\footnotesize
    \begin{tabular}{l| l | rrr | rrr | r}
    \toprule
    \bf Dataset &\bf Approach & \multicolumn{3}{c|}{\bf AIS} & \multicolumn{3}{c|}{\bf Rouge Prec.} & \bf Length \\
     &  & Strict & Macro & Micro & R1 & R2 & RL & \# Tokens \\
    \midrule
    \midrule
    \multirow{2}{*}{Researcher} & Plan-based Retrieval (Var.B) &  \multirow{1}{*}{81.76} & \multirow{1}{*}{93.29} & \multirow{1}{*}{93.81} & \multirow{1}{*}{98.12} & \multirow{1}{*}{86.86} & \multirow{1}{*}{97.84} & \multirow{1}{*}{104.62} \\
     & w/o unanswerable & 79.25 & 92.49 & 93.01 & 98.24 & 86.77 & 97.88 & 108.28 \\
    \midrule
     \multirow{2}{*}{News Events} & Plan-based Retrieval (Var.B) & \multirow{1}{*}{82.69} & \multirow{1}{*}{95.85} & \multirow{1}{*}{96.01} & \multirow{1}{*}{99.72} & \multirow{1}{*}{93.62} & \multirow{1}{*}{99.55} & \multirow{1}{*}{135.72} \\
     & w/o unanswerable & 76.92 & 94.59 & 93.54 & 99.55 & 93.02 & 99.43 & 147.03 \\
        \bottomrule
    \end{tabular}
    \end{center}
    \caption{\textbf{Importance of Marking Questions as Unanswerable.} We measure the importance of marking questions that did not have an answer from the QA step as unanswerable (first row) as opposed to dropping the question from the context in the generation prompt (second row).
    We report these metrics on the Researcher and News Events Dataset with the \textunicorn model. Highlighting unaswerable questions leads to increase in the AIS metric.}
    \label{tab:unanswerable}
    \vspace{-5mm}
\end{table*}

\subsection{Metrics}
\label{sec:metrics}
Our goal is to measure the rate at which models generate hallucinated text about the given entities and events. We will focus on the case where hallucination detection is about having a source document that supports the generated text. (We acknowledge that this differs from the case of generating objectively true statements.) To measure whether or not a generated sentence is attributable to one of the sources used by the model to produce the generated text, we compare the following metrics:

\begin{figure}[t]
    \centering
    \includegraphics[width=0.48\textwidth]{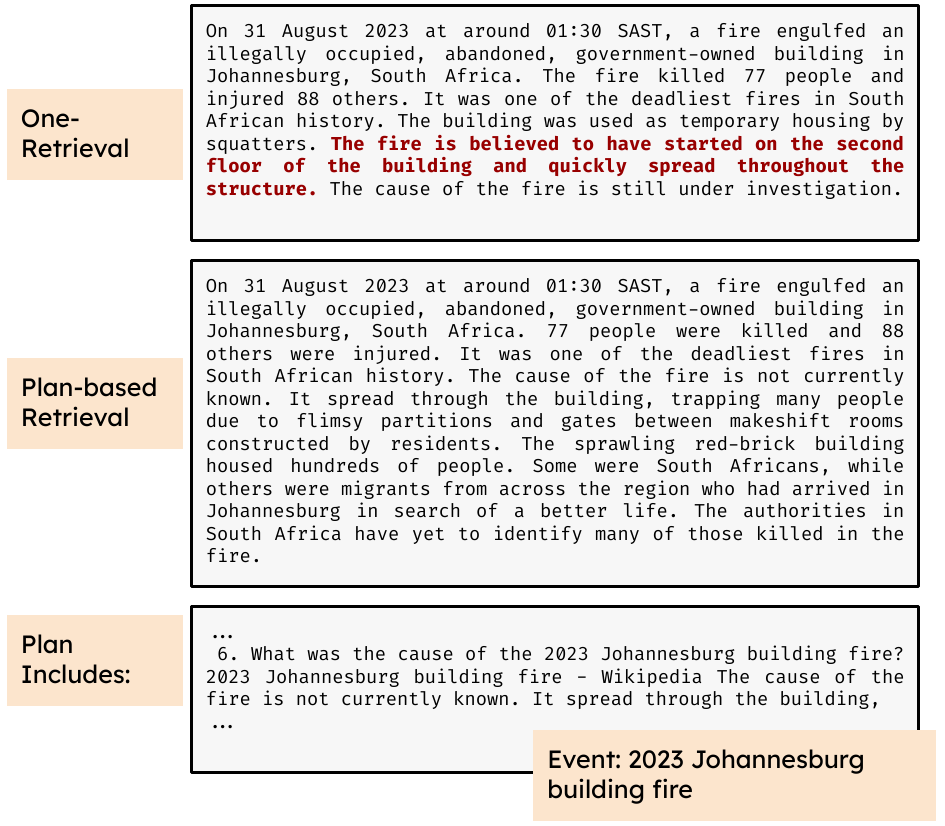}
    \caption{\textbf{Example Generation}. One of the hallucinations in the \searchOnly model is the focus of one of the questions provided in the question-based plan.}
    \label{fig:example}
    \vspace{-5mm}
\end{figure}

\textbf{Attribution.} We evaluate whether a model generation is hallucinated by checking whether it can be attributed to the source documents. Specifically, we use an automated metric that approximates the Attributable to Identified Sources (AIS) metric of \citet{rashkin2023measuring}. The AutoAIS model \citep{honovich2022true}\footnote{\href{https://huggingface.co/google/t5\_xxl\_true\_nli\_mixture}{https://huggingface.co/google/t5\_xxl\_true\_nli\_mixture}} is trained on several natural language inference (NLI) datasets and predicts whether a given context passage supports a claim. We segment the language models' output into sentences and evaluate whether a sentence can be attributed to any of the context passages (retrieved snippets and QA model answers). We report 3 variants of the metric based on aggregation. Strict AIS measures what fraction of the model outputs have all sentences correctly attributed. Macro AIS reports what fraction of the output sentences per query can be attributed on average. Micro AIS reports what fraction of all output sentences across all prompts are attributable to the context. For \paramOnly, we check attribution to web search snippets obtained with the original query.

\textbf{Rouge Prec.} We also measure the overlap between the generated text and the evidence sources using ROUGE precision metrics for ROUGE-1 (R1), ROUGE-2 (R2) and ROUGE-Lsum (RL) with the retrieved documents and answers as the reference. 

\textbf{Length} Since different approaches use different prompts and different amounts of context, they naturally tend to differ in terms of the length of the final LM output. We report the number of words in the output from different approaches. Measuring length is important to consider when observing the strict AIS metric since writing more means there are a larger number of required sentences to be correctly attributed (or in other words, more chances for mistakes).

\textbf{Human Evaluation} The focus of this work is to improve attributable generation. However, with a human evaluation study (see Appendix~\ref{app:human_eval}), we demonstrate that the generation quality does not degrade in fluency while being more informative.

\subsection{Empirical Results}
\label{exp:results}

We first consider a comparison of the approaches from \S~\ref{sec:method_variants},
in terms of the metrics described in \S~\ref{sec:metrics} (AIS, ROUGE, length) on each of the four datasets from \S~\ref{sec:dataset}. Table~\ref{tab:main_table_bison} presents the results in this setting using the \textbison model and Table~\ref{tab:main_table_unicorn} presents results using \textunicorn. For each metric, we report the average across three different runs with nucleus sampling.

{\bf Observation 1. Using Parametric Knowledge Alone Yields Fewer Attributed Sentences.} \paramOnly{} never generates summaries that are completely attributed (Strict AIS score is always 0). However, the non-zero Macro AIS scores suggest that the language models retain knowledge from large portions of their training corpus. In particular, we notice that the Macro AIS scores are higher for entities that had Wikipedia pages prior to the model knowledge cutoff (Wiki-Ent and Wiki-Event) than both less public (Researcher) and newer entities (News Events). Note that the more recently released \textunicorn has higher Macro AIS score in this purely parametric setting, perhaps indicating more parametric knowledge.

{\bf Observation 2. Adding Retrieved Evidence Improves Attribution Compared to using Parametric Knowledge Alone.} In comparison to \paramOnly, the retrieval-augmented approach, \searchOnly, has a much higher attribution rate. The high ROUGE scores indicate that the model is copying and stitching together facts from the search snippets. This is the expected behavior given that the model is instructed to only use information present in its context. The text-unicorn-001 appears to be better at following this instruction than text-bison-001. We see that there is headroom for reducing hallucinations in the model outputs. One possibility to improve attribution is to simply double the number of search snippets (\searchOnlyLong) provided to the model. However, we see that the gain from this approach is inconsistent and it only improves attribution in 3 out of 8 settings. Thus, appending more search results based on the initial query is not enough to improve attribution.

{\bf Observation 3. Using Planning Along with Retrieval Can Improve Attribution.} 
Plan-based Retrieval
achieves better or comparable Strict AIS score as compared to \searchOnly in 6 out of 8 settings. The impact of planning first before writing can be seen on the generation length; when using plans, the final outputs are longer and cover more aspects of the biography. The outputs are around 20 words longer across all settings.

{\bf Observation 4. Gathering Information to Answer Questions in a Second Round of Search Improves Attribution}. Recall that web search happens two times in the Plan-based Retrieval model. First, with the initial entity or event name and then second, after questions are generated independently for each paragraph. This second web search is used to gather more documents to answer the questions (\S\ref{sec:planning},  Figure~\ref{fig:planning}). 
In the experiment presented in Table \ref{tab:second_search_ablation}
we pose the question of whether this additional, per-question based retrieval leads to performance improvements over answering the questions using only the documents in the first search. Indeed we see this second search to be beneficial, we find that it produces improvements of about 5 points of Macro and Micro AIS and two points of strict AIS. Furthermore, the approach with the second search writes longer texts (134 tokens compared to 98 tokens).

{\bf Observation 5. Indicating Unanswerable Questions in the Plan Improves Attribution}. Some of the questions that are generated by the model will not necessarily be answerable even when considering the second round of search. We hypothesize that these unanswerable questions could lead to hallucinations from the model, especially if text is generated to provide an answer to them. In Table~\ref{tab:unanswerable}, we compare two approaches for handling these unanswerable questions.\footnote{We determine if a question is unanswerable by confidence threshold on the QA model's response.} The first is to label unanswerable questions and provide the model with an instruction to not write about that particular question
(precisely, instead of an answer, we follow the question with the response \texttt{Not enough information. Skip this question.}).
This is the default setting and corresponds to the results presented in other tables. The second is to remove the unanswerable questions and to not include them in the in-context prompt. We see that explicitly highlighting unanswerable questions leads to more grounded generation. 

{\bf Observation 6. Effect of the Final Generation Prompt (Var.A vs. Var.B)}. In Tables~\ref{tab:main_table_bison} \& \ref{tab:main_table_unicorn}, we compare the effect of prompt format given the same evidence in the model context. In particular, we compare \planAndRetrieve (which uses the blueprint and query plan in the final generation prompt) with \planTheSearch which uses the same prompt as \searchOnly. Both methods use the same search results obtained by plan-based retrieval. We see that while Var.A improves in terms of the attribution metrics, the blueprint and query plan allow Var.B to generate longer outputs with a comparably high per-query Macro AIS score. Thus, this prompt allows for a trade-off between generation length and attribution. However, we note that this trend may depend on the underlying LLM model family (see \S~\ref{sec:mistral_7b}).

\begin{table}[]
\footnotesize
    \begin{center}
    \begin{tabular}{@{}l@{}r@{}r@{}r@{}r@{}r@{}r@{}}
    \toprule
     & \multicolumn{3}{c}{\bf AIS} & \multicolumn{2}{c}{\bf Rouge P.} & \bf Len \\
        & \ Strict & \ Macro & \ Micro & R2 & RL & \# Tok \\
    \midrule
    \midrule
     Plan-based (Var.A) & \ 87.18 & \ 96.86 & \ 96.68 & \ 94.05 & \ 99.62 & \ 100.91 \\
      \phantom{a} w/o plan & \ 68.59 & \ 86.38 & \ 84.15 & \ 85.45 & \ 97.51 & \ 92.68\\
     \midrule
     \searchOnly & \ 67.31 & \ 90.27 & \ 88.21 & \ 83.48 & \ 94.52 & \ 96.42 \\
        \bottomrule
    \end{tabular}
    \end{center}
    \caption{\textbf{Importance of Using Outline for Question Generation (on News Events).}
We compare different methods of retrieval of information while using the same final generation prompt with \textunicorn. We see that using the chain-of-thought style paragraph outline helps to produce more grounded responses than an approach that simply generates questions from the initial search results.}
\label{tab:outlingImportance}
\end{table}

{\bf Observation 7. Generating an Outline Before Generating Questions Improves Attribution}. In Table~\ref{tab:outlingImportance}, we study how the query-writing method affects the attribution while holding the generation prompt constant. In particular, we ablate \planTheSearch by skipping the outline generation step. 
The approach labeled `(w/o plan)' directly generates search queries based on the initial search results without first generating a paragraph outline. All 3 approaches use the same prompt during generation.
We see that it is not the specific generation prompt but the outline-based retrieved information procured by Plan-Based Retrieval that leads to improvements in attribution.
The significantly lower attribution rate of the `(w/o plan)' ablation compared to Plan-Based Retrieval shows that chain-of-thought style blueprint generation improves the utility of subsequently generated queries.

\textbf{Observation 8. Document set used for AutoAIS evaluation of the \searchOnly Minimally Affects Results}. One might hypothesize that the improvement in AutoAIS score seen by 
Plan-based Retrieval
is due to its having more evidence documents for attribution. We compare the AutoAIS score for the \searchOnly model with two different sets of evidence documents. In Table~\ref{tab:aisSetChange}, we see that the AutoAIS increases (from 67.31 strict AIS to 69.23 strict AIS), but is significantly lower than the 82.69 strict AIS results from 
Plan-based Retrieval.

\begin{table}[]
\begin{center}\footnotesize
    \begin{tabular}{@{}l@{}r@{}r@{}r@{}r@{}r@{}r@{}}
    \toprule
    \bf Evidence  & \multicolumn{3}{c}{\bf AIS} & \multicolumn{3}{c}{\bf Rouge Prec.} \\
         & Strict & Macro & Micro & R1 & R2 & RL \\ 
     \midrule
     \multicolumn{7}{c}{\searchOnly} \\
     \midrule
     Normal & \quad  67.31 & \quad  90.27 & \quad  88.21 & \quad  95.83 & \quad  83.48 & \quad  94.52 \\
    Expanded & \quad  69.23 & \quad  91.78 & \quad  90.45 & \quad  98.71 &  \quad 88.48 & \quad  98.22 \\
     \midrule
     \multicolumn{7}{c}{\planAndRetrieve} \\
    \midrule
    Expanded & 82.69 & 95.85 & 96.01 & 99.72 & 93.62 & 99.55 \\
    \bottomrule
    \vspace{-4mm}
    \end{tabular}
    \end{center}
    \caption{\textbf{Effect of Evidence Set on AIS Score (on News Events).} We observe that the set of evidences used for AIS evaluation does not dramatically change the evaluation metrics for the \searchOnly model. Results reported on the News Events dataset with \textunicorn. \emph{Normal} evidence refers to the \searchOnly models retrieved results. \emph{Expanded} refers to the union of the retrieved results of \searchOnly and \planAndRetrieve models.}
    \label{tab:aisSetChange}
\end{table}

\subsection{Generalization to Open-source Models}
\label{sec:mistral_7b}

\begin{table*}[t]
\begin{center}
\footnotesize
\begin{tabular}{ l | ccc | ccc | c }
\toprule
\bf Approach & \multicolumn{3}{c|}{\bf AIS} & \multicolumn{3}{c|}{\bf Rouge Prec.} & \bf Length \\
& Strict & Macro & Micro & R1 & R2 & RL & \# Tokens \\
\midrule
\midrule
\searchOnly & 11.7 & 70.5 & 71.1 & 80.7 & 47.8 & 77.6 & 168.0 \\
\searchOnlyLong & 20.3 & 76.6 & 76.6 & 87.9 & 57.3 & 84.9 & 169.6 \\
\planTheSearch & 16.0 & 73.5 & 74.0 & 91.9 & 61.1 & 89.6 &  \textbf{177.6} \\
\planAndRetrieve & \textbf{25.0} & 76.7 & \textbf{77.1} & \textbf{92.8} & \textbf{64.2} & \textbf{90.8} & 161.8\\
\bottomrule
\end{tabular}
\end{center}
\caption{\textbf{Comparison of Generation Approaches using \texttt{Mistral-7B-Instruct-v0.3}.} We see that \planAndRetrieve leads to the generation of more attributable text than the \searchOnly and \searchOnlyLong baselines methods. In this setting, \planTheSearch improves over \searchOnly but not \searchOnlyLong. These results demonstrate that plan-based retrieval with model-specific tuning is a useful approach for grounded, long-form generation.}
\label{tab:mistral_7b}
\end{table*}

To demonstrate that the studied approach of plan-based generation extends to model families not discussed in the main experiments of the paper, we conduct additional experiments with \texttt{Mistral-7B-Instruct-v0.3}~\citep{jiang2023mistral}, an open-weight model released by Mistral AI. We replicate the Wiki-Ent setting and prompt the model to generate biographies for entities on Wikipedia. We use the instruction-tuned \texttt{Mistral-7B-Instruct-v0.3} model with the same prompts as listed in \ref{app:prompts} with minor changes (chat-style formatting) suggested in the model card. Note that the absolute metric values are not comparable to the results in the main body of the paper since behaviour of search engines is non-reproducible \citep{chen2023complex}.

\textbf{Plan Quality.} From the sample plans in Tables~\ref{tab:mistral_sample_plans1},\ref{tab:mistral_sample_plans2}, we see that the model is capable of generating outlines for the output and writing search queries. Thus, this is not a capability of the specific PALM-2 models studied in the main body of the paper. Moreover, if models are unable to write plans, the PALM-2 models studied can be used to generate synthetic training data to fine-tune plan generation models.

\textbf{Attribution Quality.} Our results (see Table~\ref{tab:mistral_7b}) demonstrate that \planTheSearch and \planAndRetrieve allows the model to write biographies at a higher attribution rate than \searchOnly. Thus, the benefits of planning can be extended to other language models so long as they possess sufficient instruction following capabilities. The \searchOnlyLong baseline however performs better than the \planTheSearch approach (even when both \planTheSearch and \planAndRetrieve use the same queries and additional retrieved documents). This highlights the potential sensitivity of \texttt{Mistral-7B-Instruct-v0.3} to the formatting of the search results. Thus, the same plan-based variants may not all perform consistently across models. Still, with appropriate model-specific tuning, plan-based retrieval can improve over a single search round and is an important method to include in the system-design toolbox.

\section{Conclusion}
\vspace{-1mm}
This paper explores the use of planning to improve retrieval for grounded long-form generation. We investigate how planning can guide retrieval allowing LLMs to generate more attributable responses and provide detailed analysis of the effect of the granularity of the plans and the content and quantity of retrieved information. We show empirically how the planning-based approaches outperform, in terms of automated groundedness metrics, standard retrieval-augmented generation approaches. Our empirical investigations span topics in the top-end and long-tail of the parametric knowledge of LLMs.

\section{Limitations}

Understanding the precise causes for hallucinations in language models is beyond the scope of our work. Instead, we focus on analyzing the role retrieval has in improving attribution. While we attempt to cover both operational ranges (queries within and outside of model parametric knowledge), future work would need further verification of how retrieval augmented models perform on other domains and tasks, with other language models, and with other retrieval systems. Concurrent work by \citet{shao2024assisting-storm} demonstrates the utility of outline-based search and generation for long-form expository writing of new Wikipedia pages.

Our goal was not to compare models/retrievers, but rather analyze when and how retrieval can reduce hallucinations. Thus, we focus on evaluating the effect of retrieval scheme on one model class in different domains. We report a smaller scale evaluation of \texttt{Mistral-7B-Instruct-v0.3}~\citep{jiang2023mistral} with Plan-based Retrieval in \S~\ref{sec:mistral_7b}. We see that the popular open-weight instruction-tuned model is capable of generating paragraph outlines and subsequent search queries. We demonstrate that Plan-based Retrieval effectively improves the rate of grounded generation.

Plan-based Retrieval makes multiple calls to the LLM: (1) for outline generation, (2) for question generation, (3) for final response generation. While this may slow down the generation process, we were not proposing a finalized or productionized system. Our results provide an avenue for future research on distilling the plan writing capabilities into smaller LLMs to improve the scalability of our analyzed method. Moreover, this may be an acceptable trade-off if we can achieve more grounded long-form responses. Several past approaches have relied on multiple LLM calls for improving output quality along some metric of choice: retrieval on demand during generation \citep{jiang-etal-2023-active,trivedi-etal-2023-interleaving}, post-hoc revision of model generations \citep{gao2023rarr}, multi-sample consistency for reasoning \citep{yao2023tree}, inter alia.

Our attribution evaluation tool is a model-based one, Auto-AIS. These models are not perfect and themselves a focus of active research. Thus, the reported numbers, while effective for model rankings should not be taken at absolute face-value. See \citet{min2023factscore} and \citet{bohnet2022attributed} for further discussion on correlation between automated and human AIS evaluation.

\section{Broader Impact}

There are broad and far reaching capabilities of language models. Hallucinations can lead to a number of undesirable outcomes when using language models, from confusing user experiences, misunderstandings, to misinformation, and more. Our work should not be seen as a final solution to the hallucination problem, and should not be seen to fully quantify the extent to which any one model hallucinates or does not hallucinate. Our work is a research exploration, not a final solution.

\bibliography{references}

\appendix
\section{Appendix}

\subsection{Qualitative Evaluation of System Output}

In order to test whether plan-based retrieval and generation leads to degradation in other qualitative aspects of the output, we conduct additional evaluations beyond attribution quality.

\begin{table}[htb]
    \begin{center}
    \footnotesize
    \begin{tabular}{l c c c}
        \toprule
        Dataset & \thead{One \\ Retrieval} & \thead{No\\ Preference} & \thead{Plan-based \\ Retrieval} \\
        \midrule[\heavyrulewidth]
        \multicolumn{4}{c}{text-unicorn-001} \\
        \midrule
        Researcher & 0\% & 100\% & 0\% \\
        Wiki-Ent & 4\% & 92\% & 4\% \\
        Wiki-Event & 4\% & 92\% & 4\% \\
        News Events & 8\% & 84\% & 8\% \\
        \midrule
        Avg & 4 & 92 & 4 \\
        \midrule[\heavyrulewidth]
        \multicolumn{4}{c}{text-unicorn-001} \\
        \midrule
        Researcher & 8\% & 71\% & 21\% \\
        Wiki-Ent & 8\% & 88\% & 4\% \\
        Wiki-Event & 4\% & 92\% & 4\% \\
        News Events & 4\% & 92\% & 4\% \\
        \midrule
        Avg & 6 & 86 & 8 \\
        \bottomrule
    \end{tabular}
    \end{center}
    \caption{\textbf{Head-to-head Fluency Comparison.} Plan-based Retrieval and generation does not degrade the fluency of the output as compared to standard retrieval. In the most common scenario, the outputs are not distinguishable in terms of fluency.}
    \label{tab:fluency_eval}
\end{table}
\begin{table}[tb]
    \begin{center}
    \footnotesize
    \begin{tabular}{l c c c}
        \toprule
        Dataset & \thead{One \\ Retrieval} & \thead{No\\ Preference} & \thead{Plan-based \\ Retrieval} \\
        \midrule[\heavyrulewidth]
        \multicolumn{4}{c}{text-unicorn-001} \\
        \midrule
        Researcher & 24\% & 28\% & 48\% \\
        Wiki-Ent & 20\% & 44\% & 36\% \\
        Wiki-Event & 17\% & 8\% & 75\% \\
        Curr-Event & 8\% & 32\% & 60\% \\
        \midrule
        Avg & 17 & 28 & 55 \\
        \midrule[\heavyrulewidth]
        \multicolumn{4}{c}{text-unicorn-001} \\
        \midrule
        Researcher & 16\% & 24\% & 60\% \\
        Wiki-Ent & 36\% & 4\% & 60\% \\
        Wiki-Event & 20\% & 48\% & 32\% \\
        Curr-Event & 0\% & 36\% & 64\% \\
        \midrule
        Avg & 18 & 28 & 54 \\
        \bottomrule
    \end{tabular}
    \end{center}
    \caption{\textbf{Head-to-head Informativeness Comparison.} Plan-based Retrieval leads to model generations that are more informative about 55\% more times than standard retrieval. The improvements are clearer when the LLM is queried about long-tail entities and recent events.}
    \label{tab:info_eval}
\end{table}

\subsubsection{Human Evaluation}
\label{app:human_eval}
We had 3 annotators (2 authors and 1 colleague familiar with the scope of the experiments) annotate 25 responses from each dataset to provide a comparison between the two best-performing methods: our proposed Plan-based Retrieval approach and the \searchOnly baseline. The annotators are presented a pair of generations for the same entity/event prompt, and asked to compare the generations on two qualities: fluency and informativeness. Note that informativeness is independent of the truthfulness/correctness of the model response; we asked them to indicate whichever response provided more distinct, relevant facts. The source of each generation is hidden from the annotators and they can label the better generation or indicate no preference.

Regarding fluency (Table~\ref{tab:fluency_eval}), we find that Plan-based Retrieval with \texttt{text-unicorn-001} yields generations that are as or more fluent than the baseline \searchOnly response 92-100\% of times. It does so while producing 10-20\% more factual responses. In terms of informativeness (Table~\ref{tab:info_eval}), responses from our Plan-based Retrieval were more informative 55\% of the time compared to 18\% win-rate for the baseline \searchOnly. \searchOnly performs well on queries in the head of the distribution (Wiki-Ent and Wiki-Event); Plan-based Retrieval shows its effectiveness on the long-tail query sets. We observe similar trends in responses from the \texttt{text-bison-001} model.

\subsubsection{N-gram Repetitiveness}
\label{app:repetition}
Plan-based Retrieval uses the paragraph plan to generate search queries. Since queries for each paragraph are generated independently, there may be repeated queries in the final plan. To demonstrate that the final generations from Plan-based Retrieval are not repetitive, we compute the fraction of unique N-grams (based on word boundaries) out of all N-grams in the generation. If the generation was repetitive (increasing length by reiterating the same facts), then this metric would be low. From Table~\ref{tab:ngram_uniqueness}, we see that the \{1,2,3\}-gram uniqueness of texts from Plan-based Retrieval is comparable to the same metrics with the \paramOnly and \searchOnly baselines. In fact, we see that the fraction of unique trigrams in the generation with Plan-based Retrieval is not distinguishable from the \searchOnly setting and is generally higher than the fraction of unique N-grams in the \paramOnly settings (i.e., the generation incorporates newly retrieved information).

\renewcommand\theadfont{}
\begin{table}[tb]
\begin{center}
\scriptsize
\begin{tabular}{l m{18ex} ccc }
\toprule
Dataset & Approach & \thead{1-gram\\Unique.} & \thead{2-gram\\Unique.} & \thead{3-gram\\Unique.} \\
\midrule[\heavyrulewidth]
\multicolumn{5}{c}{text-unicorn-001} \\
\midrule
\multirow{3}{8ex}{Researcher} & \paramOnly & 71.90 & 95.13 & 98.81 \\
  & \searchOnly & 73.26 & 95.63 & 98.74 \\
  & Plan-based Retrieval\-(Var.A) & 75.23 & 96.13 & 98.89 \\
\midrule
\multirow{3}{8ex}{Wiki-Ent} & \paramOnly & 72.79 & 95.25 & 98.95 \\
  & \searchOnly & 75.57 & 96.35 & 99.18 \\
  & Plan-based Retrieval\-(Var.A) & 77.93 & 96.74 & 99.29 \\
\midrule
\multirow{3}{8ex}{Wiki-Event} & \paramOnly &  60.75 & 87.72 & 94.65 \\
  & \searchOnly &  69.33 & 93.78 & 98.12 \\
  & Plan-based Retrieval\-(Var.A) &  67.89 & 93.57 & 98.19 \\
\midrule
\multirow{3}{8ex}{News Events} & \paramOnly & 60.95 & 87.54 & 94.62 \\
  & \searchOnly & 73.30 & 95.14 & 98.29 \\
  & Plan-based Retrieval\-(Var.A) & 66.53 & 92.13 & 97.37 \\
\midrule[\heavyrulewidth]
\multicolumn{5}{c}{text-bison-001} \\
\midrule
\multirow{3}{8ex}{Researcher} & \paramOnly & 73.34 & 95.70 & 99.02 \\
  & \searchOnly & 76.02 & 96.57 & 99.17 \\
  & Plan-based Retrieval\-(Var.A) & 73.11 & 95.76 & 98.69 \\
\midrule
\multirow{3}{8ex}{Wiki-Ent} & \paramOnly & 71.18 & 94.54 & 98.70 \\
  & \searchOnly & 78.91 & 97.28 & 99.39 \\
  & Plan-based Retrieval\-(Var.A) & 74.20 & 95.97 & 99.03 \\
\midrule
\multirow{3}{8ex}{Wiki-Event} & \paramOnly &  68.82 & 93.79 & 98.46 \\
  & \searchOnly &  73.65 & 95.41 & 98.67 \\
  & Plan-based Retrieval\-(Var.A) &  71.08 & 94.25 & 98.04 \\
\midrule
\multirow{3}{8ex}{News Events} & \paramOnly & 66.80 & 92.08 & 97.43 \\
  & \searchOnly & 75.25 & 95.96 & 98.90 \\
  & Plan-based Retrieval\-(Var.A) & 70.87 & 94.86 & 98.74 \\
\bottomrule
\end{tabular}
\end{center}
\caption{\textbf{Comparison of percentage of unique n-grams in model output.} We see no noticeable indication of repetition in the outputs of Plan-based Retrieval. Further human evaluation (Tab.~\ref{tab:fluency_eval}) confirms that the output of Plan-based Retrieval is fluent.}
\label{tab:ngram_uniqueness}
\end{table}

\begin{table}[tb]
\begin{center}
\footnotesize
\begin{tabular}{p{0.48\textwidth}}
\toprule
Paragraph 1\\
\midrule

\begin{itemize}[nolistsep,noitemsep,leftmargin=*]
\item Begin by introducing Ramesses IV, the third Pharaoh of the Twentieth Dynasty of the New Kingdom of Ancient Egypt, with a mention of his original name, Heqamaatre-Setepenamun.
\item Provide the dates of his reign, which were from 1155 BCE to 1149 BCE.
\item Mention that he was the second son of Ramesses III, and likely named after Ramesses, the second son of Ramesses II.
\item Emphasize that he was appointed crown prince by the 22nd year of his father’s reign, following the deaths of his four older brothers.
\end{itemize}\\
\midrule
Paragraph 2 \\
\midrule
\begin{itemize}[nolistsep,noitemsep,leftmargin=*]
\item Describe the events leading to his ascension to the throne, such as the assassination of his father, Ramesses III, in 1156 BC, which made Ramesses IV the king at a middle-aged age.
\item Mention that his reign was relatively short, lasting only six years.
\end{itemize}\\
\midrule
Paragraph 3 \\
\midrule
\begin{itemize}[nolistsep,noitemsep,leftmargin=*]
\item Discuss Ramesses IV's burial details, noting that he was originally buried in tomb KV2 in the Valley of the Kings, West Thebes, but was later reburied in the tomb of Amenhotep II (KV35).
\item Highlight that his mummy was clumsily rewrapped and labeled by later priests.
\end{itemize}\\
\midrule
...\\
\bottomrule
\end{tabular}
\end{center}
\caption{Partial plan generated zero-shot by Mistral-7B-Instruct-v0.3 for query: Ramesses IV}
\label{tab:mistral_sample_plans1}
\end{table}

\begin{table}[tb]
\begin{center}
\footnotesize
\begin{tabular}{p{0.48\textwidth}}
\toprule
Paragraph 1\\
\midrule

\begin{itemize}[nolistsep,noitemsep,leftmargin=*]
\item Instructions: Begin by introducing Ronaldo, focusing on his full name, nationality, and the sport he is famous for. Mention his nickname, if it is not already clear from his name. Provide a brief background on his hometown and birth year.
\item Do not include information about Cristiano Ronaldo, another famous athlete with a similar name.
\end{itemize}\\
\midrule
Paragraph 2 \\
\midrule
\begin{itemize}[nolistsep,noitemsep,leftmargin=*]
\item Instructions: Highlight Ronaldo's achievements and accolades early in his career, such as winning the FIFA World Cup with Brazil in 2002 and being named the tournament's top scorer (Golden Shoe award). Mention any individual awards he received during this period, such as the FIFA Player of the Year awards in 1996–97 and 2002.
\item Do not repeat information about specific teams he played for at this stage, as it will be covered in a separate paragraph.
\end{itemize}\\
\midrule
Paragraph 3 \\
\midrule
\begin{itemize}[nolistsep,noitemsep,leftmargin=*]
\item Instructions: Describe Ronaldo's successful stint with European soccer teams, focusing on key clubs and the impact he made during his time there. Mention any notable accomplishments or awards he received during this period.
\item Do not include personal details about his life outside of soccer at this stage.
\end{itemize}\\
\midrule
...\\
\bottomrule
\end{tabular}
\end{center}
\caption{Partial plan generated zero-shot by Mistral-7B-Instruct-v0.3 for query: Ronaldo (Brazilian footballer)}
\label{tab:mistral_sample_plans2}
\end{table}

\subsection{Prompt Formats}
\label{app:prompts}

\begin{enumerate}[leftmargin=*]
\item Direct Generation Prompt
\begin{quote}
{{\footnotesize
    \tt{Write a fluent, clear paragraph about \{entity\}}.
}}
\end{quote}


\item Search-Based Generation Prompt
\begin{quote}
{{\footnotesize
    \tt Search Results: \\
    Snippet Title: \{result 1 title\}  \\
    Snippet Text: \{result 1 text\}  \\
    Snippet Title: \{result 2 title\}  \\
    Snippet Text: \{result 2 text\}  \\
    ...  \\
    Snippet Title: \{result K title\}  \\
    Snippet Text: \{result K text\}  \\
    Write a fluent, clear paragraph about \{entity\}   using only facts in the given text.
    }}
\end{quote}

\item Outline Prompt
\begin{quote}
{{\footnotesize
    \tt
    Given the above search results, write a list of instructions for how to provide an answer to write a bio about \{entity\} in the format:

    Paragraph 1: Instructions for paragraph 1
    
    Paragraph 2: Instructions for paragraph 2
    
    ...
    
    Paragraph N: Instructions for paragraph N
    
    The outline should allow for each paragraph to be written independently in parallel. The collection of paragraphs should form a bio for \{entity\}. For each paragraph, write a descriptive set of instructions for the content that should be included and summarize the things that should not be included because they are written in other paragraphs. All facts, dates, and years must be supported in the given search results.
}}
\end{quote}

\item Question Prompt
\begin{quote}
{{\footnotesize
    \tt Search Results: \\
    Snippet Title: \{result 1 title\}  \\
    Snippet Text: \{result 1 text\}  \\
    Snippet Title: \{result 2 title\}  \\
    Snippet Text: \{result 2 text\}  \\
    ...  \\
    Snippet Title: \{result K title\}  \\
    Snippet Text: \{result K text\}  \\ \\
    Given the above search results, what are the questions you would want answered to write the following paragraph \{paragraph description\} about \{entity\}? 
    Write just the questions separated by a new line. Each question should be understandable independently.
}}
\end{quote}

\item Outline-based Generation Prompt
\begin{quote}
{{\footnotesize
    \tt
    Consider the following question-answer pairs:\\
    \{Question 1\} \{Answer 1\} \\
    ... \\
    \{Question N\} \{Answer N\} \\
     Write a fluent, clear paragraph about \{entity\} using only facts in the above.
}}
\end{quote}

\end{enumerate}

\subsection{Extended Related Work}
\label{app:ext_related_work}

\noindent\textbf{Responsiveness to Context.} When provided with factual knowledge in-context, we hope that LLMs use the facts and do not hallucinate. Previous studies have considered this for small models in the presence of producing LM facts \citep{petroni2019language}. \citet{neeman2022disentqa} study how predictions change when facts in context contradict parametric knowledge. \citet{longpre2021entity,li2023kaft} performs similar analysis in the adversarial setting. \citet{yoran2023making} show that retrieving irrelevant information can hurt LLM performance. Concurrent with our work, \citet{vu2023freshllms} demonstrate that including auxiliary data from the search engine and ordering the search results such that the most relevant results are presented at the end improves LLM grounding. We conduct a similar analysis of what and how to present search results to LLMs for long-form generation.

\noindent\textbf{Understanding Hallucinations.}
The manner in which language models hallucinate is investigated both in dataset construction and empirical comparisons \cite{cao2021hallucinated,das2023diving,li2023halueval,rawte2023troubling,zheng2023does}. For instance, \citet{li2023halueval} investigates hallucinate rates as a function of model as well as task and breaks down hallucination types into error categories (e.g., ``comprehension'', ``factual'', ``specificity'', and ``inference''). Similarly, \citet{sadat2023delucionqa} presents analyze hallucinations in specific domains, in particular, car manuals. 

\noindent\textbf{Grounded Generation.} Prior work has demonstrated that generating text that is grounded in given sources (be it tabular data, or natural language) can be challenging \citep[inter alia]{wiseman2017challenges,cao2018faithful}. Methods have used constrained decoding \citep{lu2020neurologic,lu2021neurologic} to produce text that is copied from given sources. Other methods have used architectural changes, which use attention-based re-weighting to make grounding more likely \citep{choi2021may}. AIS evaluation models have also been used during training \citep{zablotskaia2023calibrating} and decoding \citep{roit2023factually,wan2023faithfulness} to increase faithfulness of model generations.

\noindent\textbf{Attribution by Self-Verification.} As discussed before, it is well known that language models store a large amount of knowledge in their parameters \citep{petroni2019language}. 
Additionally, instruction-tuned LLMs have demonstrated the ability to perform natural language inference with zero-shot prompting~\citep{wei2022finetuned}.
\citet{dhuliawala2023chainofverification} use this ability for muilt-answer QA to verify generated responses with the same LLM.
\citet{ji-etal-2023-towards} apply a similar strategy for medical domain generative-QA and show that iterative verification and patching (fixing factually incorrect statements) leads to more factually correct responses.

\noindent\textbf{In-Context Learning.} The method of providing task-specific examples in the context of language models to solve a tasks, i.e., in-context learning \citep{brown2020language}, is similar to our work in that it requires being responsive to the context provided to the language model.
There exists parallels between our work and prior methods that retrieve in-context examples \citep{liu2022makes}.

\end{document}